\def\year{2020}\relax
\documentclass[letterpaper]{article}
\usepackage{aaai20}
\usepackage{times}
\usepackage{helvet}
\usepackage{courier}
\usepackage{url}
\usepackage{graphicx}

\usepackage{microtype}
\usepackage{subfigure}
\usepackage{booktabs} % for professional tables
\usepackage{mathpazo} % Use the Palatino font by default
\usepackage{amsmath,amssymb,amsfonts}
\usepackage{tikz}
\usepackage{pgfplots}
\usepackage{siunitx}
\usepackage{tabularx}
\usepackage{comment}
\usepackage{stackengine}
\usepackage{eso-pic}% http://ctan.org/pkg/eso-pic
\usepackage{multirow}

\newcommand{\shiftleft}[2]{\makebox[0pt][r]{\makebox[#1][l]{#2}}}

\nocopyright
\title{Modeling Electrical Motor Dynamics using Encoder-Decoder with Recurrent Skip Connection}
\author{
Sagar Verma,\textsuperscript{\rm 1}\textsuperscript{,}\textsuperscript{\rm 2}*
Nicolas Henwood,\textsuperscript{\rm 2}
Marc Castella,\textsuperscript{\rm 3}
Francois Malrait,\textsuperscript{\rm 2} \\
\bf \Large Jean-Christophe Pesquet,\textsuperscript{\rm 1} \\
\textsuperscript{\rm 1} Universit\'{e} Paris-Saclay, CentraleSup\'{e}lec, Inria, Centre de Vision Num\'{e}rique \\
\textsuperscript{\rm 2} Schneider Toshiba Inverter Europe \\
\textsuperscript{\rm 3} Samovar, CNRS, T\'{e}l\'{e}com SudParis, Institut Polytechnique de Paris\\
\{sagar.verma, jean-christophe.pesquet\}@centralesupelec.fr, marc.castella@telecom-sudparis.eu, \\
\{sagar.verma, nicolas.henwood, francois.malrait\}@se.com}

\begin{document}

\maketitle
\footnote{*Sagar Verma is the corresponding author. Homepage: \url{https://sagarverma.github.io/}.}

\begin{abstract}
Electrical motors are the most important source of mechanical energy in the industrial world. Their modeling traditionally relies on a physics-based approach, which aims at taking their complex internal dynamics into account. In this paper, we explore the feasibility of modeling the dynamics of an electrical motor  by following a data-driven approach, which uses only its inputs and outputs and does not make any assumption on its internal behaviour. We propose a novel encoder-decoder architecture which benefits from recurrent skip connections. We also propose a novel loss function that takes into account the complexity of electrical motor quantities and helps in avoiding model bias. We show that the proposed architecture can achieve a good learning performance on our high-frequency high-variance datasets. Two datasets are considered: the first one is generated using a simulator based on the physics of an induction motor and the second one is recorded from an industrial electrical motor. We benchmark our solution using variants of traditional neural networks like feedforward, convolutional, and recurrent networks. We evaluate various design choices of our architecture and compare it to the baselines. We show the domain adaptation capability of our model to learn dynamics just from simulated data by testing it on the raw sensor data. We finally show the effect of signal complexity on the proposed method ability to model temporal dynamics.
\end{abstract}

\section{Introduction}
\label{sec:intro}
Electrical motors are so much a part of everyday life that we seldom give them a second thought. For example, when we switch on an electrical vehicle, we confidently expect it to run rapidly up to the correct speed, provide acceleration, stop when brakes are applied, and casually predict faults to avoid future mishaps. Electrical motors have very complex dynamics and it is essential to have a controller that can provide robust control based on these dynamics. Electrical motor controllers also provide protection and supervision of the electro-mechanical system \cite{Sylvester1987,Siskind1978cpts}. For these services, it is imperative to know the dynamical physical model of electrical motors. Accurate dynamics is derived from the first principles of physics. These dynamical models are dependent on different electrical motor physical quantities like currents, voltages, speed, fluxes, inductances, and resistances, which are measured directly or indirectly using sensors or estimators. Accurately measuring some of these quantities is hard due to the presence of noise. Operating conditions also affect some of these quantities, one example being thermal evolution of resistances with time. Therefore mathematical models cannot be fully trusted in the design of controllers. A large number of simulations and human expert knowledge is required to develop robust controllers. The focus of this work is to model relationships between different electrical quantities of an electrical motor. We focus on deriving currents and electromagnetic torque from voltages and speed recorded from electrical motors using sensors.

Internet-of-Things (IoT) has made it possible to monitor different electro-mechanical devices in real-time and also provides sensor data that can be used to learn the dynamics of the system under consideration. End-to-end learning of temporal dynamics from time-series data has been made easier due to methods like Convolutional Neural Network (CNN), Recurrent Neural Network (RNN), and Long-Short Term Memory (LSTM) structures. By providing a large amount of multidimensional data, it has been shown that RNN and LSTM approaches can model complex nonlinear feature interactions which are crucial to model complex nonlinear dynamics.

In line with the current work of \cite{miller2018stable},
we found that one dimensional CNNs provide better results. We modified CNNs into an encoder-decoder architecture and incorporated recurrent skip connections between corresponding layers of the encoder-decoder architecture. In order to reduce the number of parameters, we further modified the recurrent architecture by diagonalizing its weights. We compared our proposed model with different benchmarks using a novel metric which takes into account the complexity of the predicted signals. We also proposed a novel loss function that takes into account the complexity of the signal while training the networks. We showed how using the proposed loss function for training leads to better generalization.

This paper makes the following contributions:
\begin{itemize}
    \item This is one of the first works addressing the problem of learning nonlinear dynamics of electrical motors from recorded time-series data.
    \item We propose a new Encoder-Decoder architecture to learn time-series relationship between different electrical quantities.
    \item We validate our methodology on two datasets; a large dataset of simulated electrical motor operations and a small dataset of sensor data recorded from the real-world operations of electrical motors.
    \item We propose a novel loss function that uses fast variations present in the electrical motor signals to avoid model bias.
    \item We analyse the capability of the proposed method by using a new analysis technique and we demonstrate the transfer learning capability of our approach.
\end{itemize}

The paper is structured as follows: Section \ref{sec:back} provides a brief background on electrical motors and recent advances in physics and time-series modeling. Section \ref{sec:data} describes the data and more specifically how it is collected and preprocessed. Section \ref{sec:modeling} describes the benchmark and the proposed methods. Sections \ref{sec:expers} and \ref{sec:results} contains experimental details and the obtained results. In the last section, conclusions are drawn and some possible extensions of this work are mentioned.

\section{Background}
\label{sec:back}
\begin{figure*}[ht!]
    \centering
    \stackunder[6pt]{\includegraphics[scale=0.51]{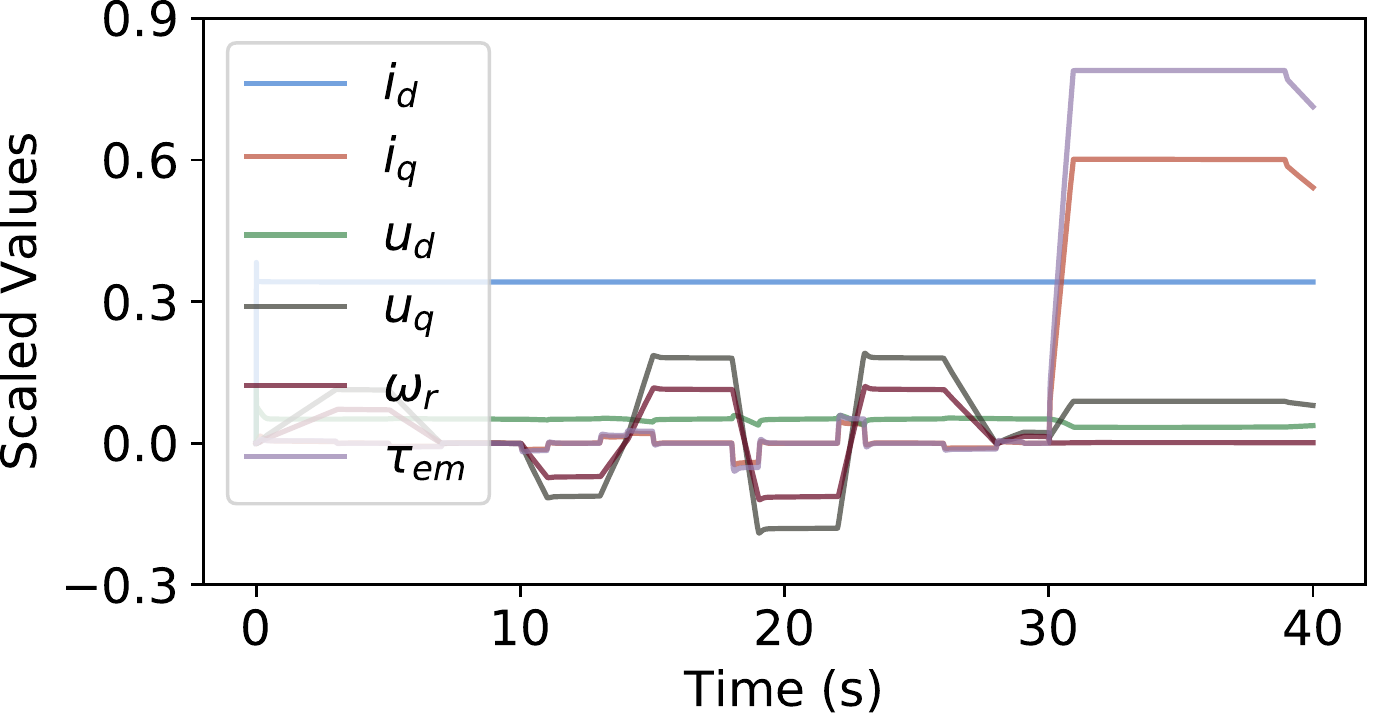}}{(a) Simulated sample.}
    \stackunder[6pt]{\includegraphics[scale=0.51]{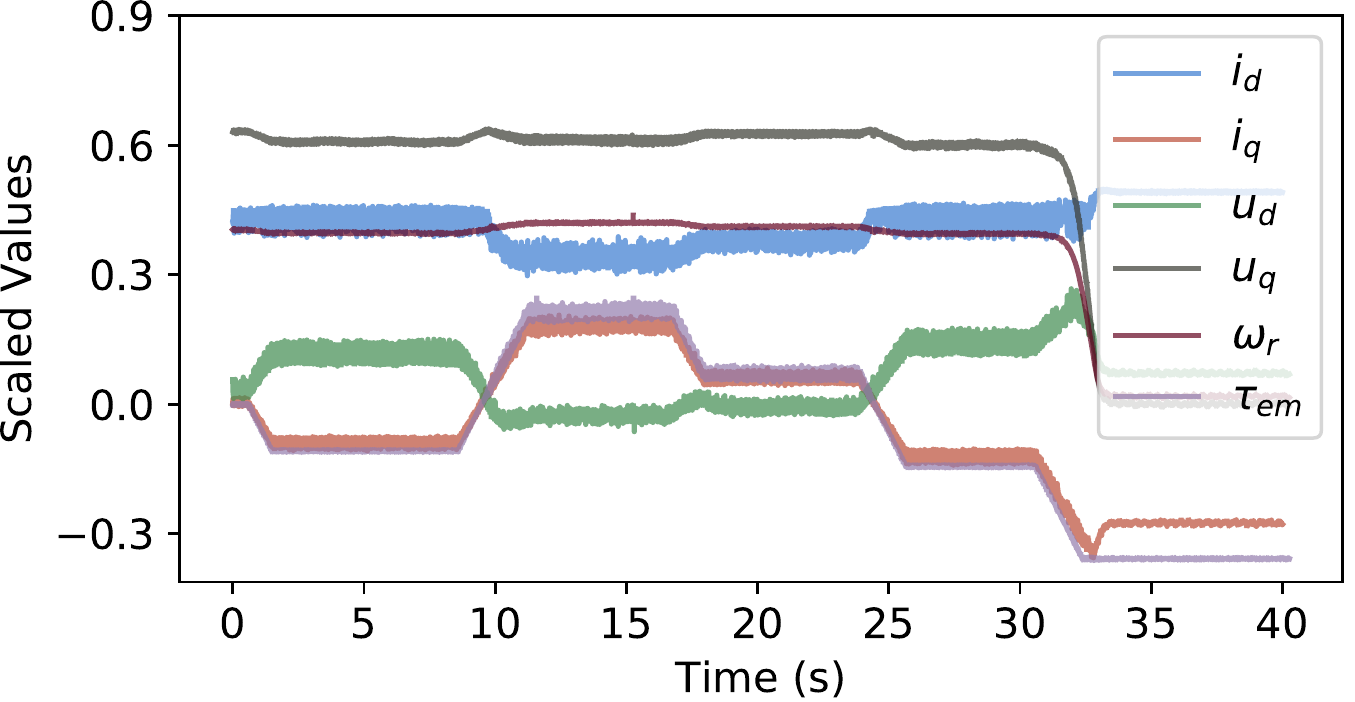}}{(b) Real world sample.}
    \caption{Electrical motor quantities showing first 40 seconds of an electrical motor operation. }
    \label{fig:sim_raw_data}
\end{figure*}

The state space model of an induction motor is presented in \cite{jadot2009induct}. Modeling of electrical motors based on analytical mechanics and energy consumption is presented in \cite{jebai2014cdc}. Existing methods for designing a controller for an induction motor can be done in two ways; when perfect knowledge of the parameters is available \cite{espinoisa1994scl,espinosa1995tac,nicklasson1997tac} and when there is an uncertainty associated with the parameters estimation \cite{chan1990tie,stephan1992iasam,marino1999tac,marino2000tac}. Electrical quantities like resistances and inductances are roughly estimated in most of the applications. These quantities also vary with change in temperature of the electrical motor environment. The control law is dependent on these quantities and measuring them requires high precision sensors and numerous expensive experiments. Due to this, acquiring perfect knowledge of the parameters is very impractical and has very limited applicability in industrial settings. Designing controllers in the presence of parametric uncertainty is done by using adaptation schemes. Two methods of adaptation are time-scale separation and time-varying adaptation \cite{anderson1977tac,zhang2002tac,jadot2009induct}. \cite{silva2013fault} presents a neural network classifier for fault diagnosis in electrical motor operations. They do not use dynamics modeling and only rely on supervised labels \cite{murphey2006model}, learn motor dynamics from simulated data and perform fault detection in simulated motor operations.

The first use of neural networks to model physical phenomena was presented in \cite{levin1991nips}. This paper presents a multi-layered neural networks for nonlinear prediction and system modeling from time-series data. Recently, deep neural networks have been used in learning physical dynamics from data in range of applications e.g.,  calorimetry \cite{carminati2017nipsw}, drone landing \cite{shi2018neurallander}, and nonlinear dynamics identification \cite{lusch2017deep}. Karpatne et al. presents a physics-guided neural network (PGNN) that leverages the output of physics-based model simulations along with observational features to generate predictions using a neural network architecture \cite{karpatne2017nipsw}. Furthermore, they present a novel framework for using physics-based loss functions in the learning objective of neural networks, in order to ensure that the model predictions not only show lower errors on the training set but are also scientifically consistent with the known physics on the unlabeled set.

\begin{table*}[h]
    \centering
    \begin{tabular}{c c c c c}
        \toprule
          \multirow{2}{*}{\textbf{Model}} & \multicolumn{2}{c}{\textbf{Architecture}} & \multirow{2}{*}{\textbf{Input}} & \multirow{2}{*}{\textbf{Output}} \\
          \cmidrule(lr){2-3}
          & \textbf{Shallow} & \textbf{Deep} & & \\
         \midrule
         \textbf{Feed-Forward} & 4 Linear & 5 Linear & Flattened vector & Middle Value \\
         \textbf{RNN} & 1 Recurrent $\rightarrow$ 2 Linear & 2 Recurrent $\rightarrow$ 2 Linear & Channelized & Same length as input \\
         \textbf{LSTM} & 1 LSTM $\rightarrow$ 2 Linear & 2 LSTM $\rightarrow$ 2 Linear & Channelized & Same length as input \\
         \textbf{CNN} & 3 Conv $\rightarrow$ 2 Linear & 4 Conv $\rightarrow$ 2 Linear & Channelized & Middle Value \\
         \bottomrule
    \end{tabular}
    \caption{Architectural details of the benchmark models.}
    \label{tab:benchmark_models}
\end{table*}

 RNN and LSTMs have been shown to be very good at learning hidden temporal dynamics from data in various applications such as wind speed forecasting \cite{ghaderi2017icml}, estimating missing measurements in time series \cite{yoon2017icml}, and consumer event forecasting \cite{laptev2017icml}. Convolutional architectures have recently been shown to be competitive on many sequence modelling tasks when compared to the de-facto standard of recurrent neural networks (RNNs), while providing computational and modeling advantages due to inherent parallelism. In \cite{bai2018arxiv}, authors provide an empirical comparison between convolutional and recurrent network in modeling time-series. Aksan et al. presents a stochastic variant of temporal convolutional network which performs better than stochastic RNNs \cite{aksan2018stcn}. Miller et al. have shown that in some cases, feed-forward networks are better in modeling temporal patterns than sequential networks \cite{miller2018stable}.

Li et al. showed that parameters in recurrent neural networks can be decreased by making neurons independent of each other \cite{li2018indrnn}. In time-series prediction, different events often have different importance. This can be achieved using asymmetric loss function which weights distinct parts of signal differently as shown in \cite{christoffersen_diebold_1997}.

\section{Available datasets}
\label{sec:data}

It seems there is no large electrical motor operations dataset available in the research community to train deep neural networks. We thus introduce two different datasets for our experiments; one dataset consists of simulations performed by using the control law proposed in \cite{jadot2009induct} and the second dataset is recorded from an industrial electrical motor. Data is collected at a sampling rate of 250Hz. We generate 100 hours of simulation data which cover a wide range of operating conditions.The dataset consists of the following electrical quantities; currents $i_{d}$ and $i_{q}$, voltages $u_{d}$ and $u_{q}$, rotor speed $\omega_r$, stator pulsation $\omega_s$, and torque $\tau_{em}$. The indices $d$ and $q$ denote three phase quantities represented in a two phase orthogonal rotating reference frame. The real electrical motor data are collected from a 4-kilowatt induction motor. Data from 10 various operating conditions are collected. Rotor speed and torque load are different in each of the runs. In total, we collected 1207 seconds of raw sensor data.

In our experiments, we split the data into four parts; training and validation parts consist of 70\% and 30\% of the simulation data, respectively. We use 20\% of the raw sensor data to fine tune the model trained on the training set of the simulated data and the rest for testing. We do not use stator pulsation $\omega_s$ in our experiments since it has a trajectory similar to $\omega_r$. To train our network, we normalize our data between $(-1,1)$ as different electrical quantities have different ranges. Normalized signals from the first 40 seconds of a simulated sample and a raw sensor sample are shown in figure \ref{fig:sim_raw_data}. It can be seen that raw sensor data has short term variations due to inherent noise present on the sensors.

\section{Modeling}
\label{sec:modeling}
\begin{figure*}[t]
    \centering
    \includegraphics[scale=0.265]{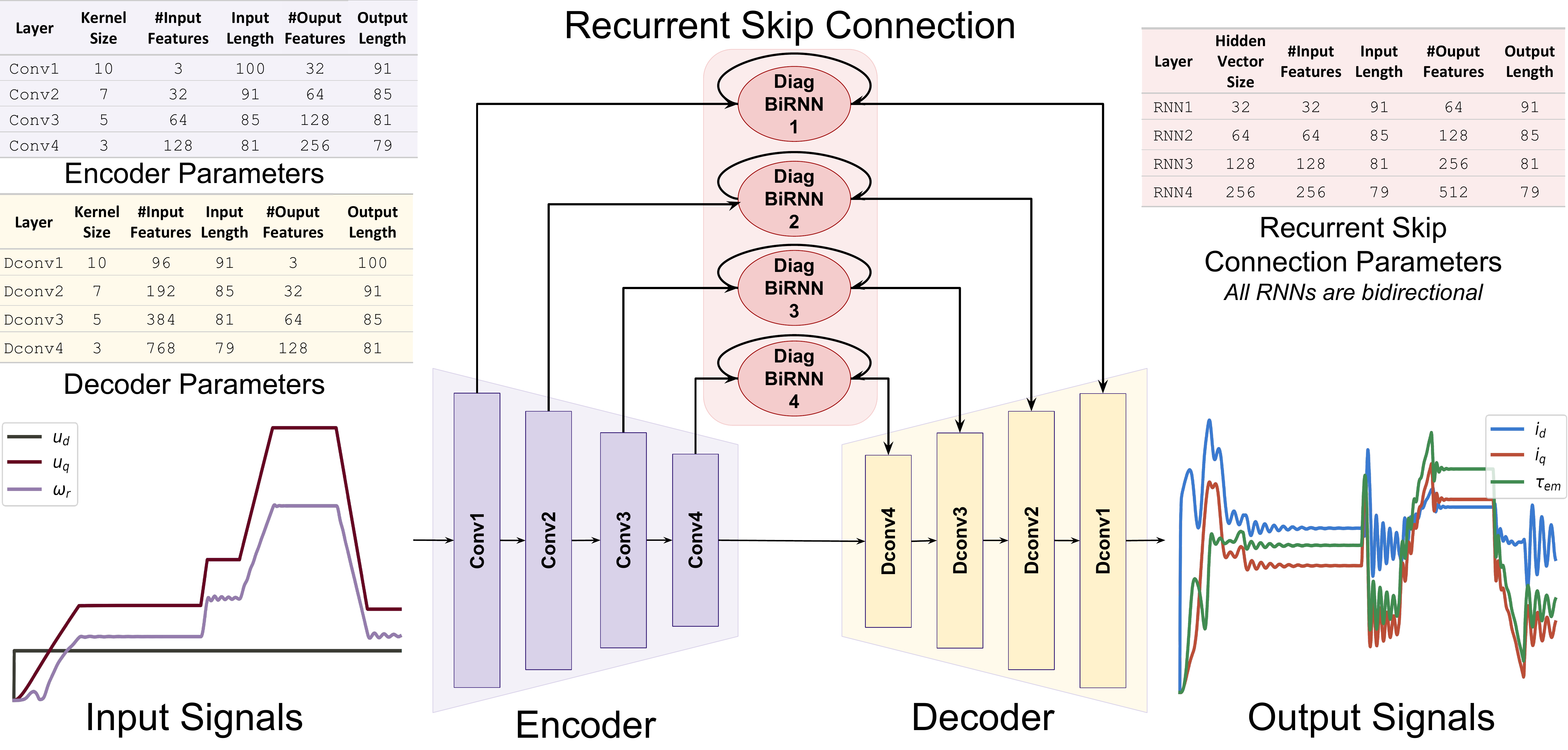}
    \caption{Proposed architecture.}
    \label{fig:arch}
\end{figure*}

\subsection{Benchmark Methods}

To the best of our knowledge, this is the first work in modeling nonlinear dynamics of electrical motor and we would also like to emphasize that the proposed datasets are challenging. To illustrate this fact, we provide several benchmark methods that are derivatives of standard neural networks.  Broadly, feed-forward network, convolutional neural network, recurrent neural network and Long-Short Term Memory (LSTM) structures are evaluated. Table \ref{tab:benchmark_models} shows all the benchmark networks. For each type of network, we try two variations, shallow and deep, to evaluate the effect of the network depth on their learning capability.

\subsubsection{Feedforward Neural Networks} We use feedfoward neural networks (FNNs) to show that our proposed problem and dataset are quite difficult and FNNs have limited learning capabilities. Row 1 in table \ref{tab:benchmark_models} provides the configuration details of the two experimented networks.

\subsubsection{Sequential Neural Networks} Sequential neural networks have been used widely to learn from sequential data. RNNs and LSTMs are two of the most commonly used sequential neural networks. Configuration details are shown in rows 2 and 3 in table \ref{tab:benchmark_models}.

\subsubsection{Convolutional Neural Networks} FNNs have very limited learning capabilities when the input data is complex like sequential or multidimensional. Recently, convolutional neural networks (CNNs) have been shown to provide competitive performances on sequential data. The configuration of benchmark for CNNs is shown in row 4 of table \ref{tab:benchmark_models}.

\subsection{Proposed Method}
Traditionally, sequential networks have been used to model temporal dynamics. In our experiments, we found that RNNs and LSTMs do not provide as good learning capability as one dimensional CNNs. Since our task is to perform multivariate prediction over the same length as the input, we use an architecture where all layers are made of convolutions. We then carefully introduce several intuitive modifications to the encoder-decoder architecture which leads to a better and parameter efficient model.

\subsubsection{Encoder-Decoder Network} To capture temporal dynamics from complete input and output window we introduce an encoder-decoder network. It consists of encoding and decoding blocks with convolutional and deconvolutional layers, respectively. The convolutional and deconvolutional blocks are followed by ReLU activations. We do not use pooling as in our experiments we found that they deteriorate the results.

\subsubsection{Encoder-Decoder Network with Skip Connection} It has been shown that adding skip connection to encoder-decoder helps in transferring high level features directly from one encoding layer to its corresponding decoding layer \cite{NIPS2016_6172}. We also introduce skip connections between encoding and decoding layers.

\subsubsection{Encoder-Decoder Network with Recurrent Skip Connection} Convolution operations are windowed over the kernel size, this means that convolution cannot learn temporal relationships which are out of the kernel sized windows. Adding recurrent layer over convolutional features can overcome this issue. This also helps in learning temporal patterns in the latent space. We add recurrent layers after every encoding layers. The output of the recurrent layer is then sent to the corresponding decoding layers.

\subsubsection{Encoder-Decoder Network with Bidirectional Recurrent Skip Connection} Bidirectional RNNs help in learning temporal patterns in both direction. For our use case, as we want to predict for each time step of the input window. Therefore, we also use bidirectional RNNs.

\subsubsection{Encoder-Decoder Network with Bidirectional Diagonalized Recurrent Skip Connection} Vanilla RNNs have a high number of parameters due to matrix multiplications between weights and features. Diagonalizing weights in the recurrent unit decreases the number of parameters.

\begin{table*}[ht!]
    \centering
    \begin{tabular}{c c c c c c}
        \toprule
          \textbf{Model} & \textbf{Window Size} & \textbf{Parameters} &  \textbf{MAE} & \textbf{SMAPE} & \boldmath{$R^2$} \\
         \midrule
         \textbf{Shallow Feed-Forward} & 25 & 751617 & 77.76 & 9.79\% & -0.59 \\
         \textbf{Deep Feed-Forward} & 20 & 1118209 & 78.91 & 8.53\% & -0.39 \\
         \textbf{Shallow RNN} & 100 & 9889 & 77.97 & 8.5\% & -0.3 \\
         \textbf{Deep RNN} & 150 & 12001 & 78.26 & 7.76\% & -0.35 \\
         \textbf{Shallow LSTM} & 50 & 13441 & 79.39 & 6.41\% & -0.26 \\
         \textbf{Deep LSTM} & 100 & 21889 & 79.58 & 6.29\% & -0.11 \\
         \textbf{Shallow CNN} & 100 & 518721 & 79.51 & 6.22\% & -0.13 \\
         \textbf{Deep CNN} & 100 & 650049 & 79.69 & 6.13\% & -0.14 \\
         \midrule
         \textbf{Shallow} & 100  & 309185 & 80.63 & 5.02\% & 0.08 \\
         \textbf{Deep} & 100  &  1096385 & 81.21 & 4.57\% & 0.29 \\
         \textbf{Skip} & 100  &  364801 & 28.96 & 3.71\% & 0.42 \\
         \textbf{RNN-Skip} & 100  &  638145 & 28.18 & 3.42\% & 0.43 \\
         \textbf{BiRNN-Skip} & 100  &  967105 & 27.96 & 3.31\% & 0.41 \\
         \textbf{DiagBiRNN-Skip} & 100  &  618465 & \textbf{26.88} & \textbf{1.09}\% & \textbf{0.95} \\
         \bottomrule
    \end{tabular}
    \caption{Experimental details and results for the benchmark models and the proposed model variants obtained on the simulated validation set. First 8 rows show results for the benchmark models and last 6 rows show results for the variants of the proposed model. Average of the three output quantities; currents $i_d$ and $i_q$ and electromagnetic torque $\tau_{em}$ is shown.}
    \label{tab:benchmark_results}
\end{table*}

The hidden state update equation of an RNN is given by:

\begin{equation}
    h_t = \text{tanh}(Wx_t + Uh_{t-1} + b)
\end{equation}
where $x_t \in \mathbb{R}^M$ and $h_t \in \mathbb{R}^N$ are the input and hidden state at time $t$, respectively. $W \in \mathbb{R}^{N \times M}$, $U \in \mathbb{R}^{N \times N}$, and $b \in \mathbb{R}^N$ are the weights for the input and the hidden vector, and the bias of the neurons. We propose to impose diagonal structures for the weight matrices $W$ and $U$ by setting $N=M$. Let the diagonal vector of entries be denoted by vectors $w$ and $u$, respectively. Practically, this amounts to replacing the matrix multiplication operations with Hadamard products $\odot$ of the involved vectors. The diagonalized recurrent network is described as:

\begin{equation}
    h_t = \text{tanh}(w \odot x_t + u \odot h_{t-1} + b)
\end{equation}
where $w \in \mathbb{R}^M$, $u \in \mathbb{R}^M$, and $b \in \mathbb{R}^M$ are input weights.

\subsection{Total Variation Weighted Mean Square Loss}
In real world usage of electrical motors, large variations in the signals occurs less often than small variations. We observe this effect in our dataset, which causes model bias toward small variations when trained with mean square loss. This is not a desirable behavior if the learned model is used in controllers. To avoid this problem, we propose a novel asymmetric loss function that takes into account the signal variations:

\begin{equation}
    \mathcal{L}_{\text{TV-WeightMSE}} = \frac{1}{N}\sum_{i=1}^{N}\sum_{t=1}^{T-1} |y^i_t - y^i_{t+1}| \frac{1}{T}\sum^{T}_{t=1}(y^i_t - \hat{y^i_t})^2
\end{equation}
where $y^i_t$ and $\hat{y^i_t}$ are the values of output and predicted sample $i$ at time-step $t$, respectively. $N$ is the number of training samples where each sample is of duration $T$.

\section{Experiments}
\label{sec:expers}
\begin{figure*}[t]
    \centering
        \stackunder[2.5pt]{\includegraphics[scale=0.29]{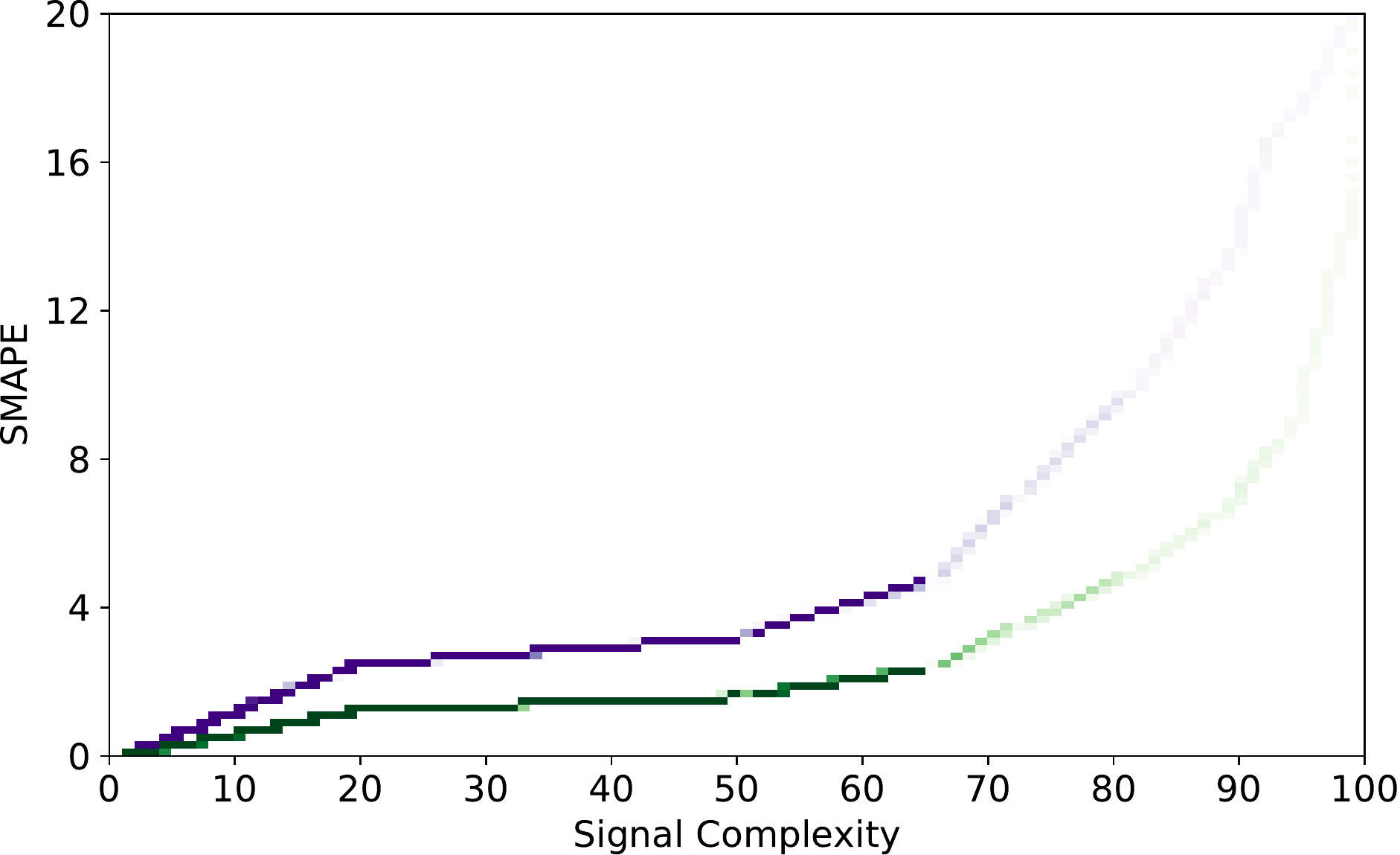}}{(a) Current $i_d$}
      \llap{\shiftleft{5.0cm}{\raisebox{2.3cm}{%  move next graphics to top right corner
      \includegraphics[scale=0.45]{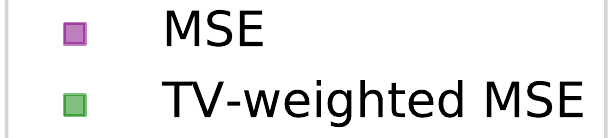}%
        }}}
      \stackunder[2.5pt]{\includegraphics[scale=0.29]{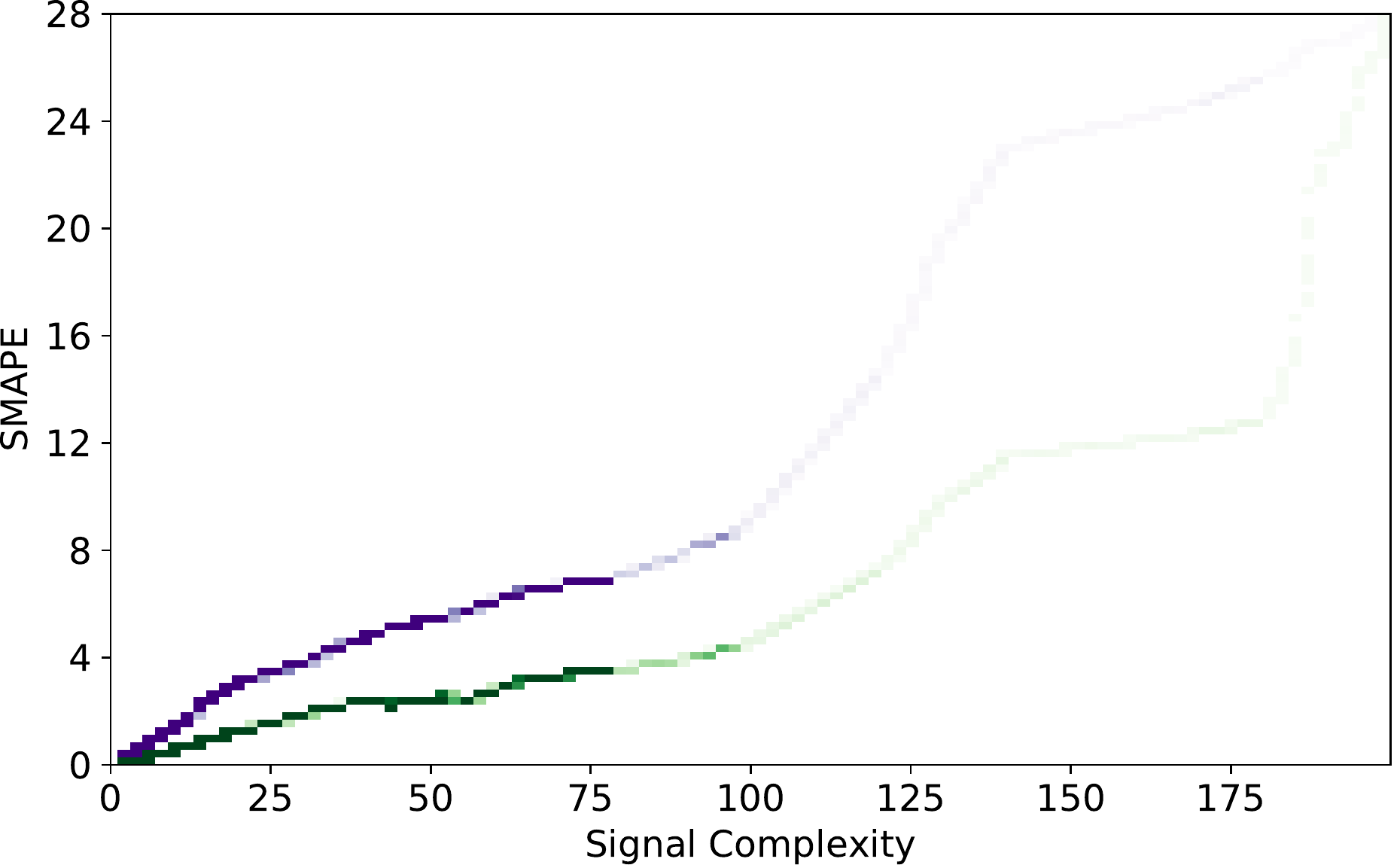}}{(b) Current $i_q$}
      \llap{\shiftleft{5.0cm}{\raisebox{2.3cm}{%  move next graphics to top right corner
      \includegraphics[scale=0.45]{tv_mse_legend.pdf}%
        }}}
      \stackunder[2.5pt]{\includegraphics[scale=0.29]{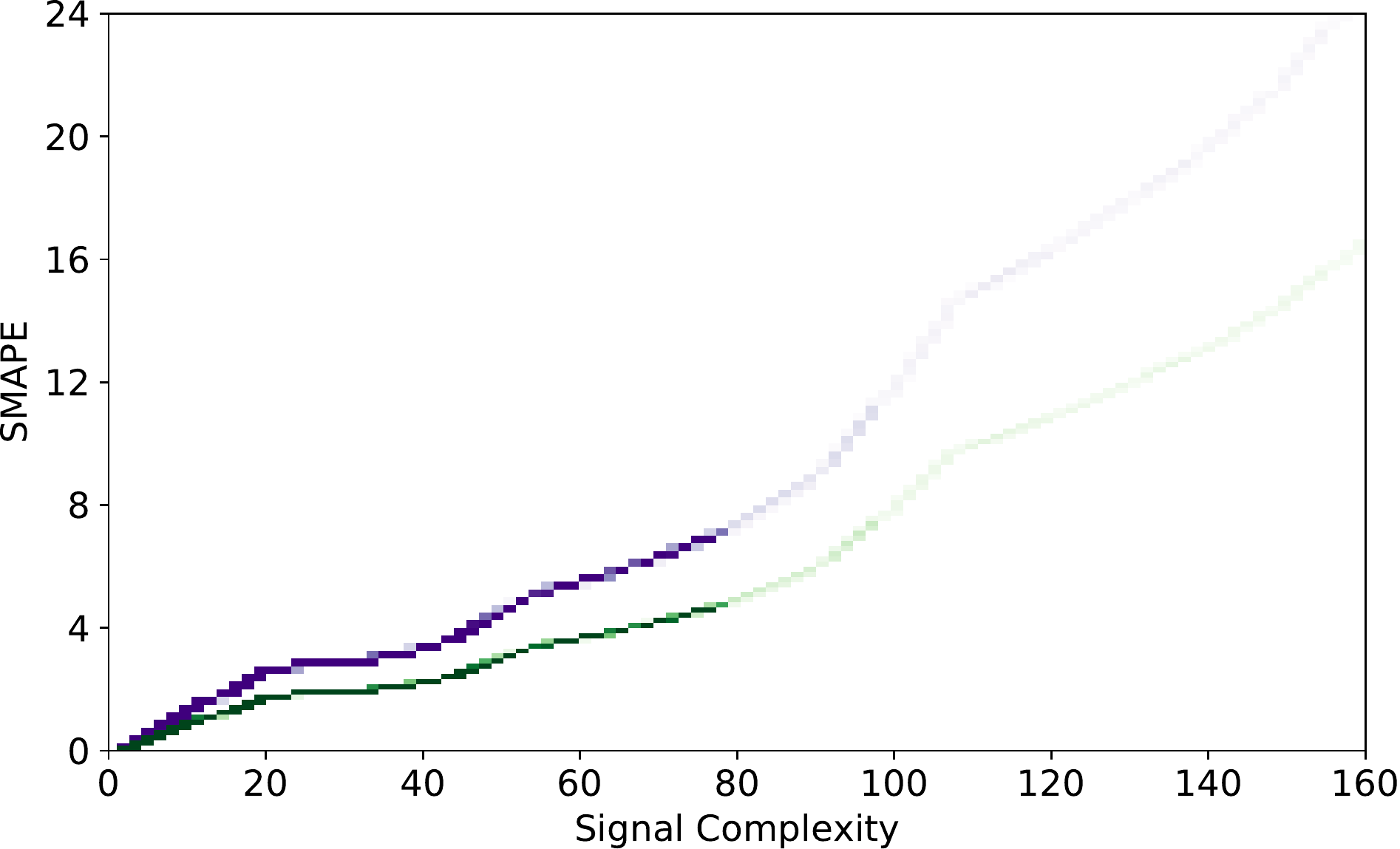}}{(c) Torque $\tau_{em}$}
       \llap{\shiftleft{5.0cm}{\raisebox{2.3cm}{%  move next graphics to top right corner
      \includegraphics[scale=0.45]{tv_mse_legend.pdf}%
        }}}
\caption{Comparison between the proposed TV-weighted MSE loss and MSE loss used to train the proposed network. TV-weighted MSE is able to learn about signal windows which have a lot of variations.}
      \label{fig:sc_smape_tv_mse}
\end{figure*}

\begin{table*}[h!]
    \centering
    \begin{tabular}{c c c c c c c}
        \toprule
         \multirow{2}{*}{\textbf{Quantity}} & \multicolumn{3}{c}{\textbf{MSE Loss}} & \multicolumn{3}{c}{\textbf{TV-weighted MSE Loss}} \\
         \cmidrule(lr){2-4} \cmidrule(lr){5-7}
         & \textbf{MAE} & \textbf{SMAPE} & \boldmath{$R^2$} & \textbf{MAE} & \textbf{SMAPE} & \boldmath{$R^2$}\\
        \midrule
        \textbf{$i_d$ (A)} & 28.1351 & 0.9702\% & 0.6572 & 27.9123 & 0.4619\% & 0.9224 \\
        \textbf{$i_q$ (A)}  & 26.8912 & 2.3925\% & 0.9507 & 26.5234 & 1.9014\% & 0.9631 \\
        \textbf{$\tau_{em}$ (Nm)} & 26.2321 & 1.5839\% & 0.9247 & 26.1978 & 0.9193\% & 0.9652 \\
        \bottomrule
    \end{tabular}
    \caption{Performance of proposed encoder-decoder with diagonalized recurrent skip connection model when trained using MSE loss and the proposed TV-weighted MSE loss. Results were obtained on the simulated data validation set.}
    \label{tab:mse_loss_results}
\end{table*}

\begin{table*}[ht!]
    \centering
    \begin{tabular}{c c c c c c c}
        \toprule
          \multirow{2}{*}{\textbf{Quantity}} & \multicolumn{3}{c}{\textbf{Simulated Model}} & \multicolumn{3}{c}{\textbf{Fine-tuned Model}} \\
          \cmidrule(lr){2-4} \cmidrule(lr){5-7}
           & \textbf{MAE} & \textbf{SMAPE} & \boldmath{$R^2$} & \textbf{MAE} & \textbf{SMAPE} & \boldmath{$R^2$} \\
         \midrule
          \textbf{$i_d$ (A)} & 39.8392 & 4.1029\% & 0.3829 & 35.3167 & 2.6429\% & 0.5637 \\
          \textbf{$i_q$ (A)} & 47.3818 & 6.3729\% & 0.4113 & 42.9472 & 5.2841\% & 0.4936 \\
          \textbf{$\tau_{em}$ (Nm)} & 38.5628 & 3.8128\% & 0.4997 & 32.3819 & 2.3891\% & 0.6017 \\
         \bottomrule
    \end{tabular}
    \caption{Results obtained for each of the output quantity on the raw test set. `Simulated Model` column shows the results of the model trained on the simulated data and `Fine-tuned Model` column shows the results of the model fine-tuned on the raw sensor data.}
    \label{tab:raw_test}
\end{table*}

For all our experiments we use an Ubuntu 18.04 OS with V100 GPU. PyTorch is employed to implement the benchmark and proposed architectures. Simulation data are collected from a Simulink model which is heavily used in the motor control industry. To show the requirements of the proposed method and why benchmark methods fail, we perform extensive experiments. We vary our architecture by trying different input lengths, number of layers, and RNN/LSTM hidden vector lengths. We try the following input lengths \{5,10,15,20,25,50,100,200\} and find out that an input length greater than 100 is better at capturing the motor operation dynamics. Depending on the architecture, different input and output structures are required. Feed-forward networks take a flattened vector and predict a single output which is the middle value of the output signal. RNNs and LSTMs take channelized input and predict output of the same length. CNNs take channelized input and predict middle value of the output signal.

In encoder-decoder variations where an RNN is used, the hidden vector size is the same as the number of features in the input vector. In the encoder-decoder network the input and output lengths are the same. We train all our models using the proposed TV-weighted mean square loss function. To find the best architecture, we use the validation set of the simulated data. Then we fine-tune the best model on the training set of the raw data and test it on the raw data test set. We also train the best performing model using mean square loss to compare it with the proposed loss function.

To evaluate the capability of the proposed method, we use different metrics allowing us to compare the performance at global and local scope of the input signal. To analyse the learning capability at global scope, we report mean absolute error (MAE), symmetric mean absolute percentage error (SMAPE), and coefficient of determination $R^2$ \cite{COLINCAMERON1997329}.

\begin{equation}
    \text{MAE}(y, \hat{y}) = \frac{1}{T}\sum^{T}_{t=1}|y_t - \hat{y_t}|
\end{equation}

\begin{equation}
    \text{SMAPE}(y, \hat{y}) = \frac{100}{T} \sum_{t=1}^T \frac{|\hat{y_t}-y_t|}{|\hat{y_t}|+|y_t|}
\end{equation}

\begin{equation}
   R^2(y,\hat{y}) = 1 - \frac{\sum^{T}_{t=1}(\hat{y_t} - \bar{y})^2}{\sum^{T}_{t=1}(y_t - \bar{y})^2}
\end{equation}
where $y_t$ is the ground truth, $\hat{y}_t$ is the predicted output of the model at time $t$, and $T$ is the total experiment duration. $\bar{y}$ denotes the mean of ground truth $y$.

MAE, SMAPE, and $R^2$ values do not provide enough information about the signal parts where the model is performing poorly or very well. Thus, we compute the signal complexity (SC) on sliding windows over the ground truth signal and plot it versus the corresponding window SMAPE value computed between the ground truth and the predicted signal. The signal complexity is given by ${\text{SC}}_y = \sum_{t=1}^{W-1} |y_t - y_{t-1}|$ for a small window length $W$. All metrics are reported on the original range of the respective quantities after re-scaling.

\section{Results}
\label{sec:results}

\begin{figure*}[h]
    \centering
      \stackunder[2.5pt]{\includegraphics[scale=0.28]{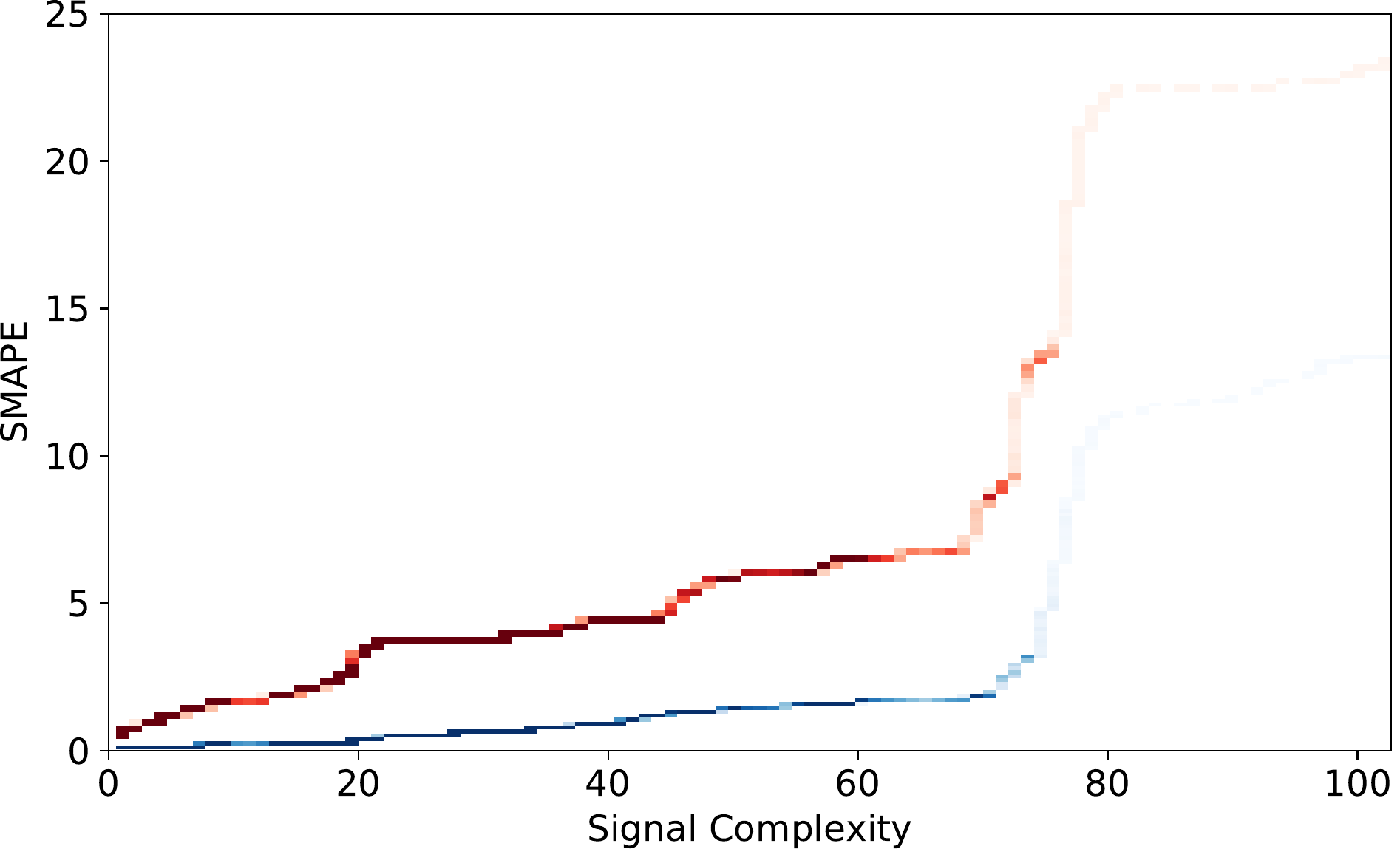}}{(a) Current $i_d$}
      \llap{\shiftleft{5cm}{\raisebox{2.3cm}{%  move next graphics to top right corner
      \includegraphics[scale=0.45]{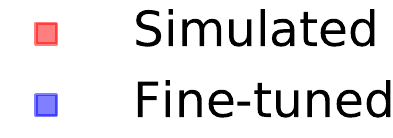}%
        }}}
      \stackunder[2.5pt]{\includegraphics[scale=0.28]{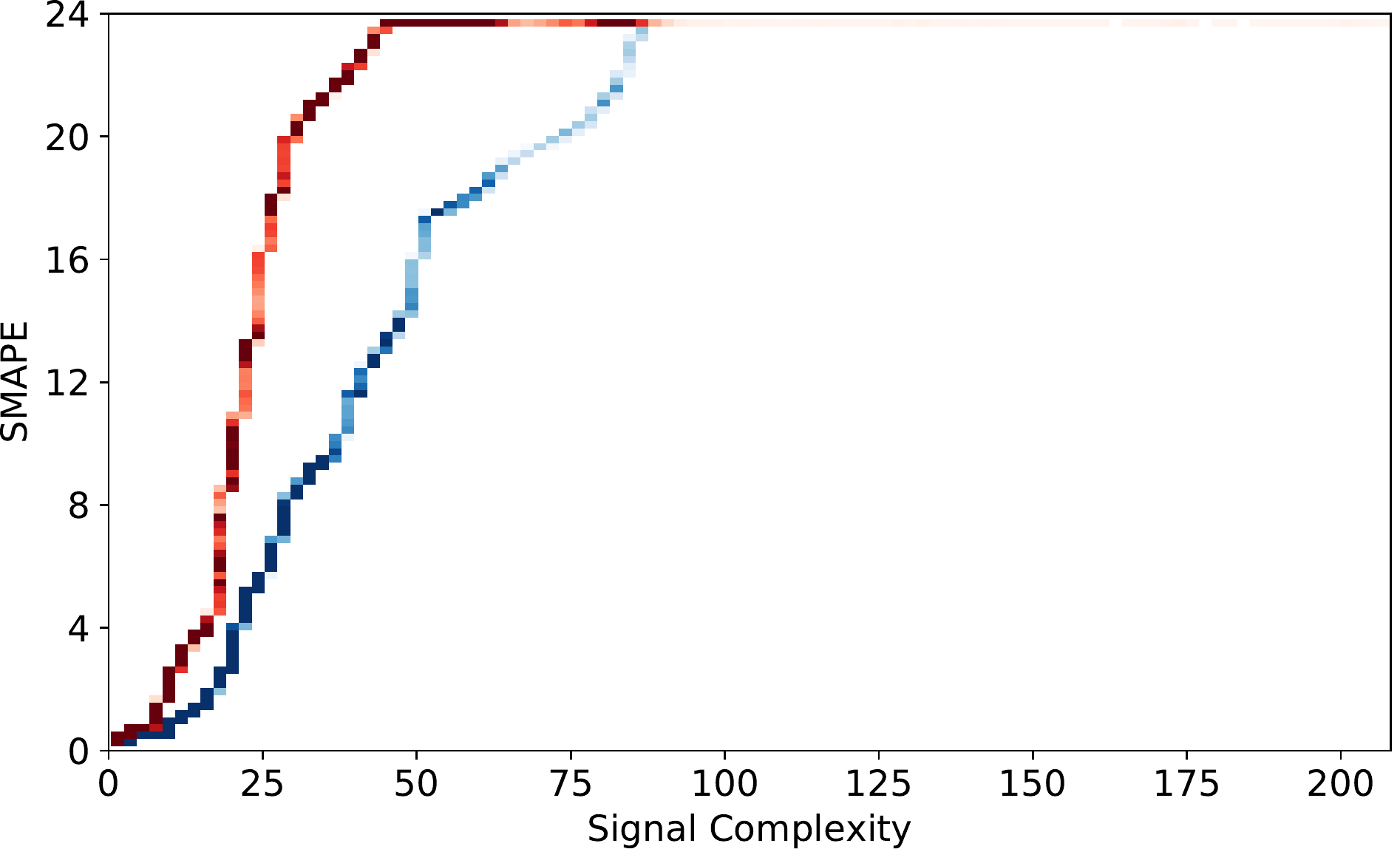}}{(b) Current $i_q$}
      \llap{\shiftleft{2.2cm}{\raisebox{0.6cm}{%  move next graphics to top right corner
      \includegraphics[scale=0.45]{legend.pdf}%
        }}}
      \stackunder[2.5pt]{\includegraphics[scale=0.28]{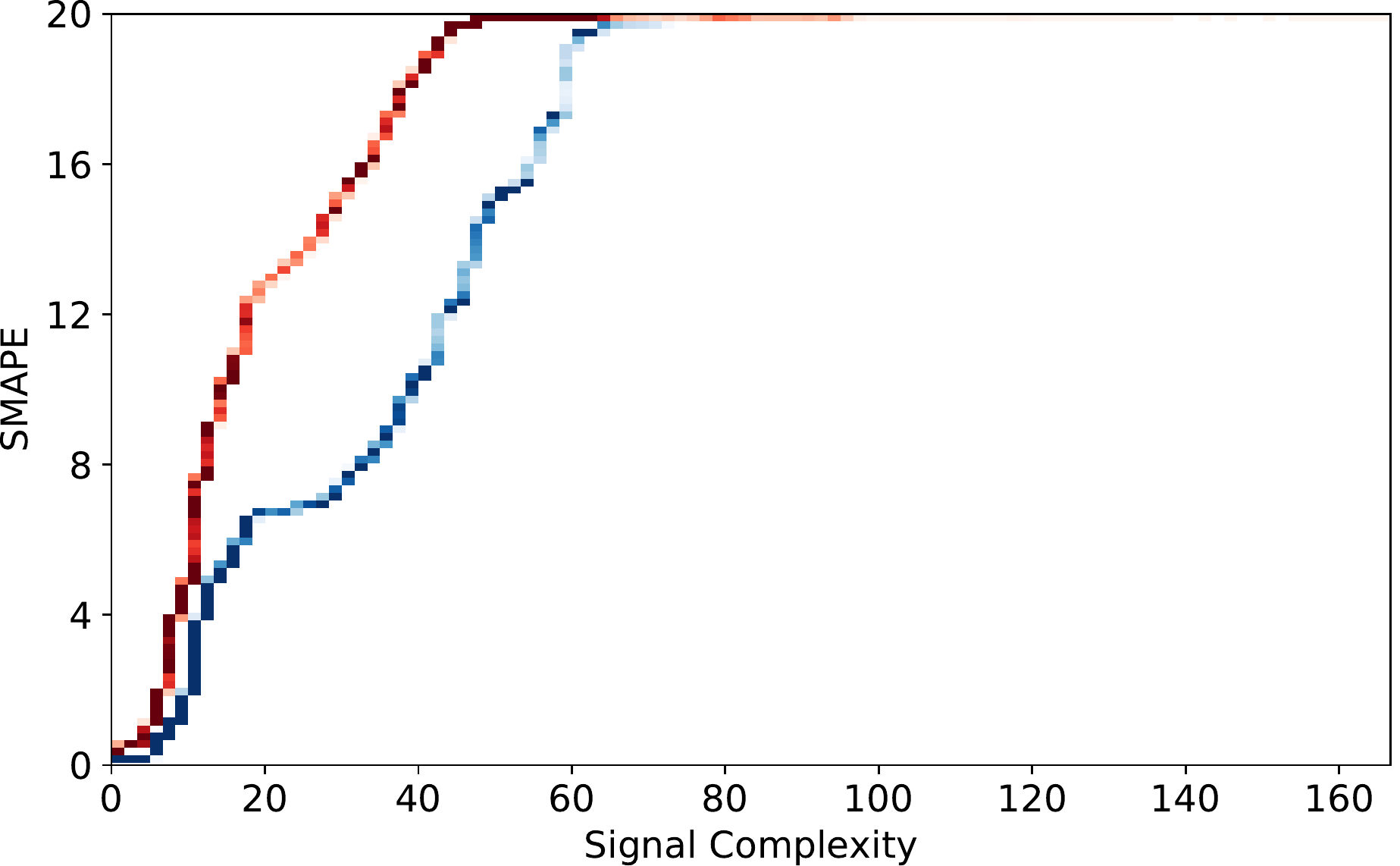}}{(c) Torque $\tau_{em}$}
       \llap{\shiftleft{2.2cm}{\raisebox{0.6cm}{%  move next graphics to top right corner
      \includegraphics[scale=0.45]{legend.pdf}%
        }}}
      \caption{Comparison of simulated and fine-tuned model using SMAPE vs Signal Complexity graph.}
      \label{fig:sc_smape}
\end{figure*}

\begin{figure*}[h!]
    \centering
      \stackunder[2.5pt]{\includegraphics[scale=0.3]{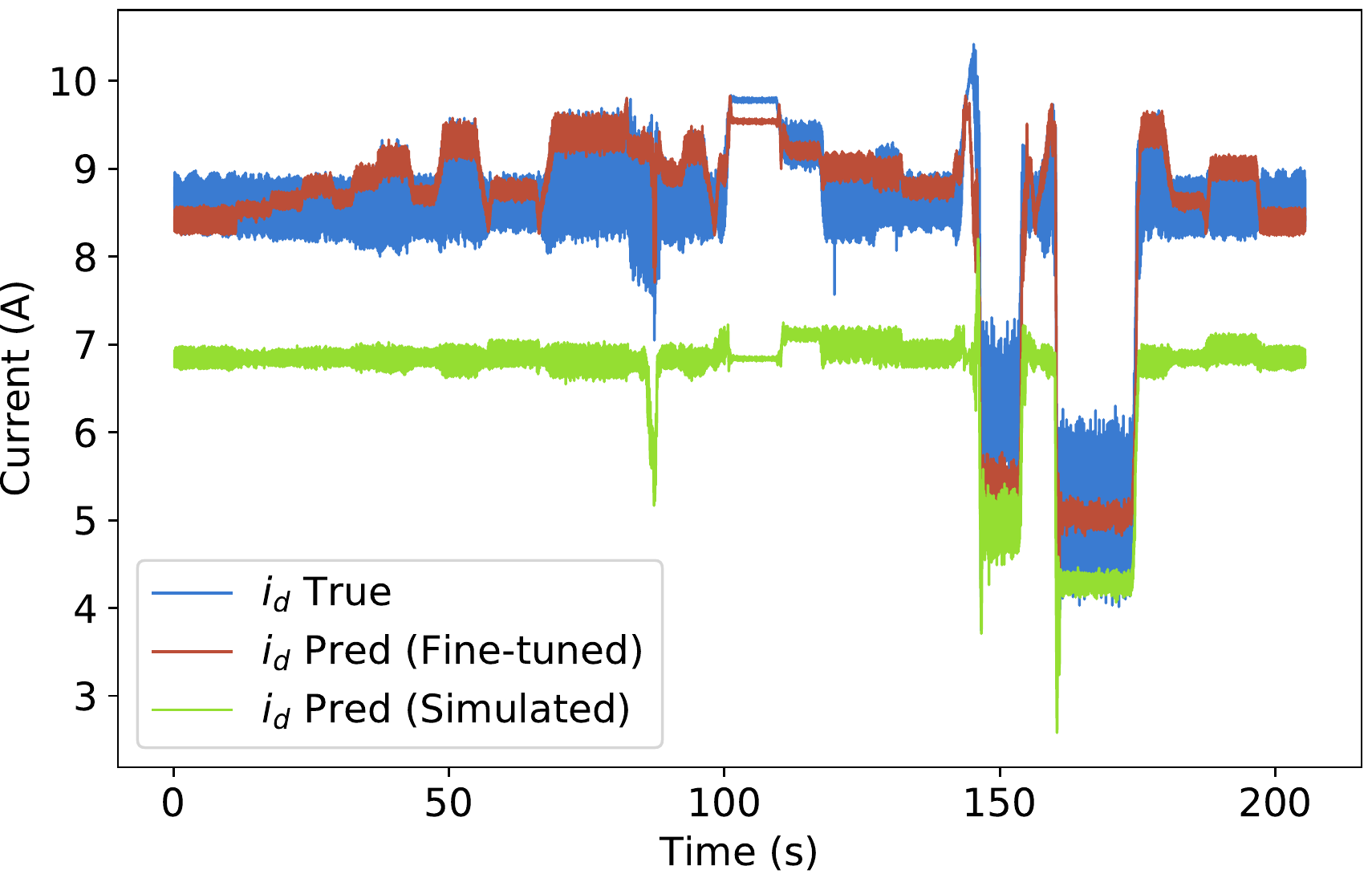}}{(a) Current $i_d$}
      \stackunder[2.5pt]{\includegraphics[scale=0.3]{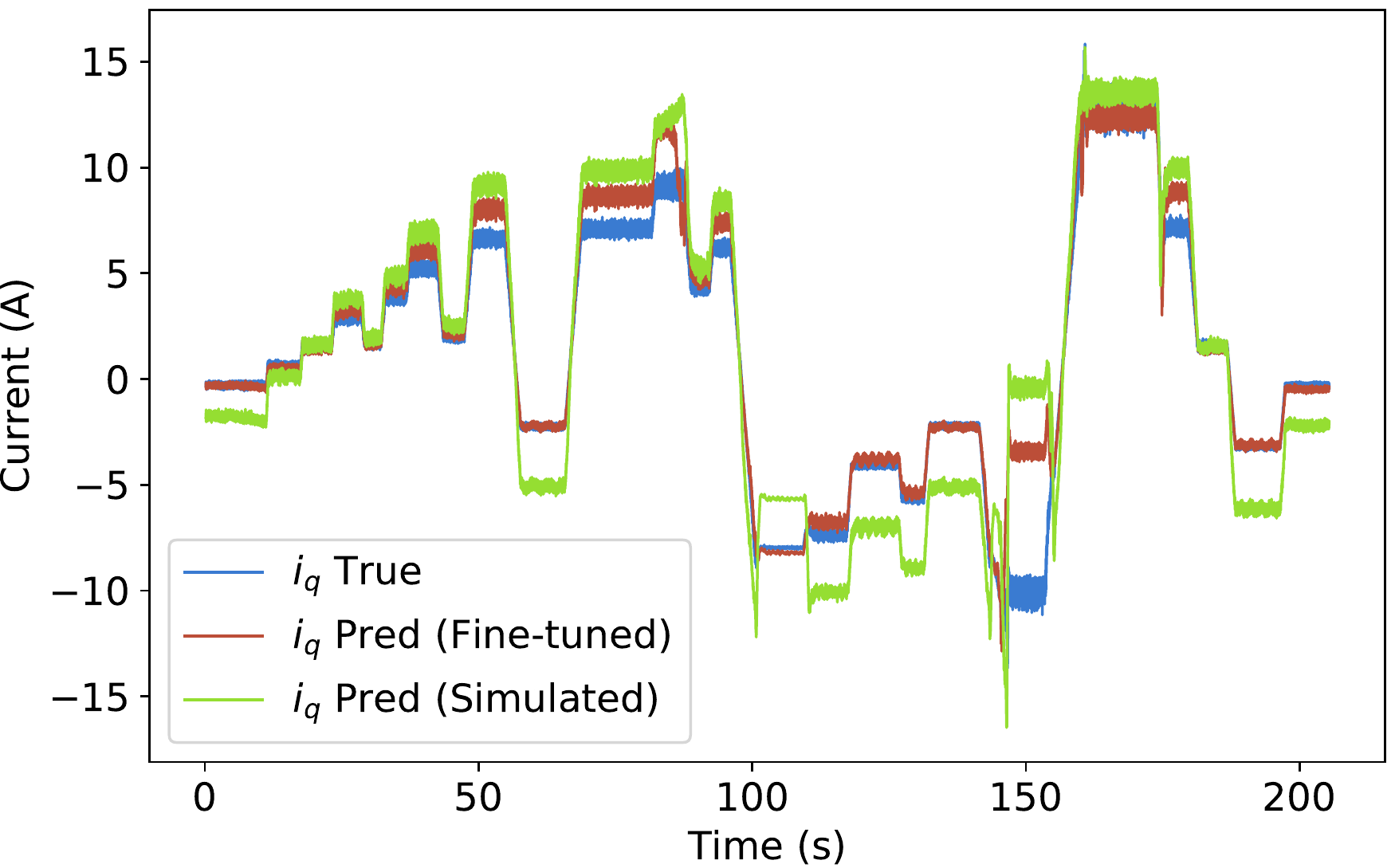}}{(b) Current $i_q$}
      \stackunder[2.5pt]{\includegraphics[scale=0.3]{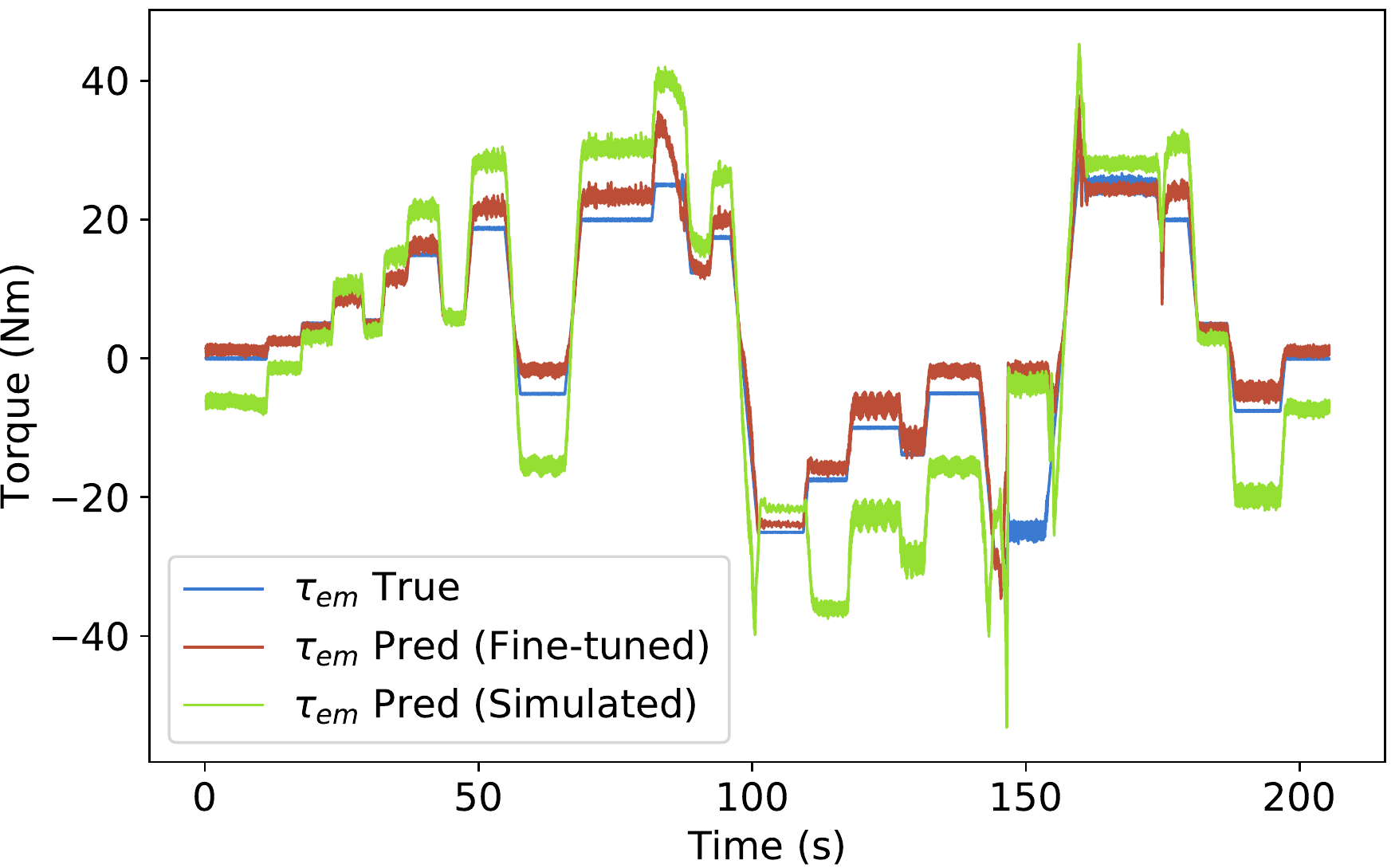}}{(c) Torque $\tau_{em}$}
      \caption{Predicted result of one of the experiments from test set.}
      \label{fig:sim_sensor}
\end{figure*}

We provide results for the benchmark models and the variations of the proposed architecture. The first eight rows of table \ref{tab:benchmark_results} show the results obtained by different benchmark models. Window size column shows the input length on which the best result was obtained. Hidden vector size for both RNN and LSTM is 32. The number of parameters is also reported for all the models. For each of them we report MAE, SMAPE, and $R^2$ values. The average of the three output quantities are provided; current $i_d$, current $i_q$, and electromagnetic torque $\tau_{em}$. All results were obtained on the validation set of the simulated data. Among benchmark models, we observe that MAE values are very close for all the models. But when we compare SMAPE and $R^2$ values, deep CNN and deep LSTM come out to be the best. In our experiments, we observe that the models perform better when the input length is 100 or more. For all the models, the performance gap between shallow and deep variants is small. This means that the network depth provides little advantage in learning nonlinear dynamics of electrical motors.

Based on the results obtained from the benchmark methods, we fix the input size to 100 for all our proposed model variants. The last 6 rows of table \ref{tab:benchmark_results} show the results of the proposed model variants trained and validated on the simulated data. First and second rows show the results of the shallow and deep variant of the encoder-decoder architecture. We see that MAE is still comparable to the benchmark models but SMAPE and $R^2$ value improves. Third row shows the result of the model where skip connections have been added between encoder-decoder. MAE gets better in this case. Fourth and fifth rows correspond to recurrent skip connections with unidirectional and bidirectional recurrence, respectively. Having recurrence in skip connections improves MAE and SMAPE values but comes at the cost of an increased number of parameters. It can be seen that bidirectionality has a positive effect on MAE and SMAPE. Last row shows the best version of our encoder-decoder model, where we replace RNNs in skip connections with diagonalized RNNs. This model outperforms all the methods and has fewer parameters when compared to other RNN variants.

Table \ref{tab:mse_loss_results} shows the results obtained by the proposed model when MSE loss and TV-weighted MSE loss were used in training. All three metrics for all three quantities improve when the proposed TV-weighted MSE loss is used in training the DiagBiRNN-Skip Encoder-Decoder network. Figure \ref{fig:sc_smape_tv_mse} shows how the SMAPE values increase when signal complexity increases, and compares the results with both loss functions. The SMAPE vs SC plots are 2D histograms where color intensity of each box represent the number of samples that are in that bin. We see that signal parts with higher signal complexity occur less often. Our model trained with MSE loss is able to predict more accurately parts of signal with small signal complexity. We observe that the model trained using TV-WeightMSE loss overcomes this issue.

Table \ref{tab:raw_test} shows the results of simulated model and model fine-tuned on the raw data training set when tested on the raw data test set. It can be seen that the proposed model is able to learn the temporal dynamics of each of the quantities very well just from the simulated data. When the model is fine-tuned on the sensor data, it seems to be able to learn about the noise associated with the sensors and yields better results. Figure \ref{fig:sc_smape} shows the SMAPE vs signal complexity graph for the three output quantities obtained from the simulated and fine-tuned models. We observe that current $i_q$ and torque $\tau_{em}$ have some signal parts which are more complex than current $i_d$. Figure \ref{fig:sim_sensor} shows the results for one of the raw samples from the test set. It can be seen that the model trained on the simulated data has some offset in its prediction whereas the model fine-tuned on the sensor data is much closer to the ground truth, even if it is still not perfect.

\section{Conclusion}
\label{sec:concl}
A novel problem has been investigated: the learning of electrical motor dynamics from time-series sensor data. We also have presented a novel encoder-decoder architecture that uses diagonalized recurrent skip connections to learn the complex temporal dynamics. To learn the model, a novel loss function has been introduced that avoids prediction bias. We have used transfer learning to fine tune a model trained on large simulated data on a small raw sensor dataset. Our experiments have shown the promising performance of the proposed method on a noisy sensor dataset collected in an industrial context. We have also carried out a detailed analysis at the global and the local scope of the prediction performed on the test data. Our results show the feasibility of AI solutions in modeling electrical motor dynamics, thus opening a new avenue of research in this area.


\bibliographystyle{aaai}
\fontsize{9.0pt}{10.0pt} \selectfont
\bibliography{ref}

\end{document}